\documentclass[letterpaper, 10 pt, conference]{ieeeconf}  

\IEEEoverridecommandlockouts                              
\usepackage[para,online,flushleft]{threeparttable}
\usepackage{textcomp}
\usepackage{booktabs}
\usepackage{multirow}
\usepackage{adjustbox}
\usepackage{graphicx}
\usepackage{siunitx}
\usepackage[utf8]{inputenc}
\let\proof\relax
\let\endproof\relax

\usepackage{amsthm}

\theoremstyle{definition}

\newtheorem{definition}{Definition}

\newtheorem{example}{Example}

\usepackage{epsfig} 
\usepackage{amsmath} 
\usepackage{amssymb}  
\usepackage{siunitx}
\usepackage{subcaption}
\usepackage{caption}
\usepackage{color}
\usepackage{theoremref}
\usepackage[noadjust]{cite}
\usepackage{amsthm}
\usepackage{mathtools}
\usepackage{algorithm}
\usepackage[noend]{algpseudocode}

\title{\LARGE\bf Optimization-based Task and Motion Planning under Signal Temporal Logic Specifications using Logic Network Flow}
\author{Xuan Lin$^{1}$, Jiming Ren$^{1}$, Samuel Coogan$^{2}$, and Ye Zhao$^{1}$ 
\thanks{$^{1}$George W. Woodruff School of Mechanical Engineering, Georgia
Institute of Technology, Atlanta, GA, 30332 USA (e-mail: \{xlin373, jren313, ye.zhao\}@gatech.edu).}
\thanks{$^{2}$School of Electrical and Computer Engineering, Georgia Institute
of Technology, Atlanta, GA 30332 USA (e-mail: sam.coogan@gatech.edu)}
\thanks{}
}

\begin{document}
\maketitle
\thispagestyle{empty}
\pagestyle{empty}

\begin{abstract}
This paper proposes an optimization-based task and motion planning framework, named ``Logic Network Flow", to integrate signal temporal logic (STL) specifications into efficient mixed-binary linear programmings. In this framework, temporal predicates are encoded as polyhedron constraints on each edge of the network flow, instead of as constraints between the nodes as in the traditional Logic Tree formulation. Synthesized with Dynamic Network Flows, Logic Network Flows render a tighter convex relaxation compared to Logic Trees derived from these STL specifications. Our formulation is evaluated on several multi-robot motion planning case studies. Empirical results demonstrate that our formulation outperforms Logic Tree formulation in terms of computation time for several planning problems. As the problem size scales up, our method still discovers better lower and upper bounds by exploring fewer number of nodes during the branch-and-bound process, although this comes at the cost of increased computational load for each node when exploring branches.

\end{abstract}
%
%

\section{Introduction}
\label{Sec:introduction}
Task and motion planning (TAMP) with temporal logic provides formal guarantees for provably correct robot plans and task completion  \cite{plaku2016motion, zhao2024survey, li2021reactive, he2015towards, shamsah2023integrated}. Particularly, TAMP with Signal Temporal Logic (STL) constraints is often formulated as an optimization problem solved by mixed-integer linear program (MILP) \cite{cardona2023mixed, sun2022multi}. However, this STL-based planning problem is theoretically intractable due to its NP-hard nature. In practice, although MILP solvers, \textit{e.g.}, via branch and bound (B\&B), can solve in a reasonable computation time, they still suffer from the worst-case (\textit{i.e.}, exponential) complexity. To take a step toward circumventing these worst-case scenarios, this study presents a novel MILP formulation by transforming STL specifications into a form of network flow to render a tighter convex relaxation for the MILP. This formulation offers a promise in improving the efficiency of the B\&B process, which can further facilitate the solve of STL-based optimization problems more efficiently. 
STL offers an expressive task-specification language for specifying a variety of temporal tasks, and provides a powerful framework to integrate discrete and continuous actions. Planning of dynamical systems under STL specifications has been widely explored for robot manipulation \cite{takano2021continuous, nawaz2024reactive}, bipedal locomotion \cite{gu2024walking, gu2024robust}, and multi-agent systems \cite{nikou2018timed, sun2022multi}. As an evaluation of the task completion, a robustness metric is introduced \cite{donze2010robust} to facilitate the search for the optimal solution. An approximation to robustness has been designed \cite{mehdipour2019average, gilpin2020smooth, pant2018fly} to entirely avoid mixed-integer programming through gradient-based optimization. This ``smooth" method improves computational speed but sacrifices the completeness of the results that is guaranteed by MILP.

\begin{figure}[t!]
    \centering
    \includegraphics[width=0.45\textwidth]{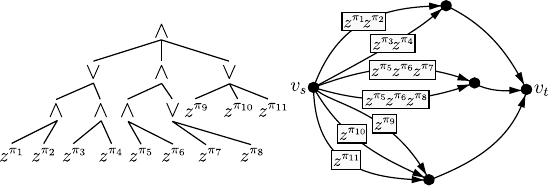}
     \caption{An example of a Logic Tree (\textit{Left}) and a Logic Network Flow (\textit{Right}) for the specification $\varphi=((z^{\pi_{1}} \wedge z^{\pi_{2}}) \vee (z^{\pi_{3}} \wedge z^{\pi_{4}})) \vee ((z^{\pi_{5}} \wedge z^{\pi_{6}}) \wedge (z^{\pi_{7}} \vee z^{\pi_{8}})) \vee (z^{\pi_{9}} \vee z^{\pi_{10}} \vee z^{\pi_{11}})$.}
\label{fig:example4}
\end{figure} 


To improve computational efficiency when solving MILP with STL constraints, existing efforts have focused on either reducing the problem size \cite{kurtz2022mixed} or tightening the convex formulation \cite{marcucci2024graphs}. In particular, \cite{marcucci2019mixed} discusses several approaches for designing MILP formulations with tight convex relaxation. In addition, \cite{marcucci2024graphs} introduces a compact formulation named Graph-of-Convex-Sets (GCS) to solve hybrid motion planning problems leveraging a structure similar to network flows, which is a classical model developed for urban traffic flow management \cite{ahuja1988network}. \cite{kurtz2022mixed} uses GCS to solve motion planning problems with a different class of temporal logic specifications, namely Linear Temporal Logic (LTL), through automata-based methods. However, building an automata is computationally expensive, and its size grows exponentially with the length of the specification \cite{wolff2014optimization}. To avoid these limitations, STL-based approaches directly encode temporal logic constraints into mixed-integer optimization.


In this paper, we reformulate the problem to achieve a tighter convex relaxation inspired by \cite{marcucci2024graphs}. Our main contributions include converting temporal logic specifications into a Logic Network Flow that encodes STL constraints. As shown in Fig. \ref{fig:example4}, temporal logic predicates are placed on the edges of a Logic Network Flow (\textit{right}) instead of on the leaf nodes of a Logic Tree (\textit{left}), which is designed in previous literatures \cite{wolff2014optimization, raman2014model}.
Our formulation is evaluated on multi-robot coordination and searching tasks. Simulation studies show that, through integrating Logic Network Flows with dynamics encoded as Dynamic Network Flows \cite{yu2013multi}, our formulation can discover tighter lower and upper bounds by exploring fewer nodes during the B\&B process compared to \cite{wolff2014optimization, raman2014model}. However, the trade-off is that solving a single node tends to be more computationally expensive. In future works, we aim to reduce the computation time on single nodes with techniques such as parallel computing.

\section{Background}
\label{Sec:background}
\subsection{Temporal Logic Preliminaries}
We consider a discrete-time nonlinear system in the form of
\begin{equation}
  \boldsymbol{x}_{t+1} = f(\boldsymbol{x}_{t}, \boldsymbol{u}_{t})  
\label{eqn:dyn}
\end{equation}
where $\boldsymbol{x}_{t} \in \mathcal{X} \subseteq \mathbb{R}^{n_x} \times \mathbb{B}^{n_z}$, $\mathbb{B} = \{0,1\}$, represents the state vector, consisting of continuous variables of size $n_x$ and binary variables of size $n_z$; $\boldsymbol{u}_{t} \in \mathcal{U} \subseteq \mathbb{R}^{n_u}$ represents the control input of size $n_u$, with $t=0,1,\ldots,T$ denoting the time indices. Given the first state of the trajectory $\boldsymbol{x}_{0} \in \mathcal{X}_0$ (where $\mathcal{X}_0$ is typically a singleton set containing only the initial condition), and the control inputs at each step, a run of the system is expressed as $\xi=(\boldsymbol{x}_{0}\boldsymbol{u}_{0})(\boldsymbol{x}_{1}\boldsymbol{u}_{1})\cdots$ via rolling out Eqn. \eqref{eqn:dyn}. We also use $\boldsymbol{x} = \{\boldsymbol{x}_0, \ldots, \boldsymbol{x}_T\}$ and $\boldsymbol{u} = \{\boldsymbol{u}_0, \ldots, \boldsymbol{u}_{T-1}\}$ to represent the state and control trajectories.

In this paper, we focus on bounded-time signal temporal logic (STL) formulas built upon convex predicates. That said, the maximum trajectory length $T$ to determine logic satisfiability  is predefined and finite. We recursively define the syntax of STL formulas as follows \cite{belta2019formal}:
$
\varphi \coloneqq \:
\pi \;|\; \neg\varphi \;|\; \varphi_1 \wedge \varphi_2 \;|\; \varphi_1 \vee \varphi_2 \;|\; \Diamond_{[t_1,t_2]}\;\varphi \;|\; \square_{[t_1,t_2]}\;\varphi \;|\; 
\varphi_1 \; \mathcal{U}_{[t_1,t_2]}\; \varphi_2
$, 
where the semantics consists of not only boolean operations “and” ($\wedge$) and “or” ($\vee$), but also temporal operators ``always" ($\square$), ``eventually" ($\lozenge$), and ``until" ($\mathcal U$). $\varphi$, $\varphi_1$, $\varphi_2$ are formulas, and $\pi$ is an atomic predicate $\mathcal{X} \rightarrow \mathbb{B}$ whose truth value is defined by the sign of the convex function $g^\pi : \mathcal{X} \rightarrow \mathbb{R}$. In this paper, we assume that the convex function is a combination of linear functions, which can be expressed as $g^\pi(t) = ({\boldsymbol{a}^{\pi}})^\top \boldsymbol{x}_t + b^\pi$. A binary predicate variable $z^\pi_t \in \mathbb{B}$ is assigned to each predicate at timestep $t$ such that:
\begin{gather}
\begin{aligned}
({\boldsymbol{a}^{\pi}})^\top \boldsymbol{x}_t + b^\pi \geq 0 &\Leftrightarrow z^\pi_t=1, \\  
({\boldsymbol{a}^{\pi}})^\top \boldsymbol{x}_t + b^\pi < 0 &\Leftrightarrow z^\pi_t=0
\label{eqn:predicate}
\end{aligned}
\end{gather}
A run $\xi$ that satisfies an STL formula $\varphi$ is denoted as $\xi \models \varphi$. The satisfaction of a formula $\varphi$ having a state signal $\boldsymbol{x}$ beginning from time $t$ is defined inductively as in Table \ref{tab:STL_satisfy}.
\begin{table}[t!]
\centering
\caption {\label{tab:STL_satisfy} Validity semantics of Signal Temporal Logic}
\vspace{-0.1in}
\begin{tabular}{l c c} \\
\hline
$(\boldsymbol{x},t) \models \varphi_1 \wedge \varphi_2$ &$\Leftrightarrow$& $(\boldsymbol{x},t) \models \varphi_1 \wedge (\boldsymbol{x},t) \models \varphi_2$ \\
$(\boldsymbol{x},t) \models \varphi_1 \vee \varphi_2$ &$\Leftrightarrow$& $(\boldsymbol{x},t) \models \varphi_1 \vee (\boldsymbol{x},t) \models \varphi_2$ \\
$(\boldsymbol{x},t) \models \Diamond_{[t_1,t_2]}\varphi$ &$\Leftrightarrow$& $\exists {t^{'}\in[t+t_1,t+t_2]}, (\boldsymbol{x},t^{'}) \models \varphi$ \\
$(\boldsymbol{x},t) \models \square_{[t_1,t_2]}\varphi$ &$\Leftrightarrow$& $\forall {t^{'}\in[t+t_1,t+t_2]}, (\boldsymbol{x},t^{'}) \models \varphi$\\
$(\boldsymbol{x},t) \models {\varphi_1}\mathcal{U}_{[t_1,t_2]}{\varphi_2}$ &$\Leftrightarrow$& $\exists {t^{'}\in[t+t_1,t+t_2]}, (\boldsymbol{x},t^{'}) \models \varphi_2$ \\
&& $\wedge \ \forall {t^{''}\in[t+t_1,t^{'}]} (\boldsymbol{x},t^{''}) \models \varphi_1$ \\ 
\hline
\end{tabular}
\end{table}

\subsection{Logic Tree}
Logic Tree \cite{wolff2014optimization, raman2014model}, also referred to as STL Tree \cite{kurtz2022mixed}, STL Parse Tree \cite{leung2023backpropagation}, and AND-OR Tree \cite{sun2022multi}, is a hierarchical data structure encapsulating STL formulas to facilitate efficient optimization solving. Here we first provide its definition and an example of translating an STL formula to a Logic Tree:

\begin{definition}
\label{Def: Logic Tree}
A Logic Tree (LT) $T^\varphi$, constructed from an STL specification $\varphi$, is defined as the tuple $(\circ, \Pi,  \mathcal{N}, \tau)$, where:
\begin{itemize}
\item $\circ \in \{\wedge, \vee\}$ denotes the combination type;
\item $\Pi = \{\pi_1, \ldots, \pi_{|\Pi|}\}$ is the set of $|\Pi|$ predicates associated with each leaf node in the tree $T^\varphi$. Each leaf node is assigned a variable $z^{\pi_i}$ to indicate its validity. 
\item $\mathcal{N}=\{T^{\varphi_0}, T^{\varphi_1}, \ldots, T^{\varphi_n}\}$ represents the set of $n+1$ internal nodes having at least one child, where the root node is denoted by $T^\varphi = T^{\varphi_0}$. Each node is associated with an STL formula $\varphi_i$ and a combination type $\circ$. Similarly, each internal node is assigned a variable $z^{\varphi_i}$ to indicate the formula's validity.
\item $\tau = \{t^{\varphi_0} , t^{\varphi_1} , \ldots , t^{\varphi_n}\}\cup \{t^{\pi_1}, \ldots, t^{\pi_{|\Pi|}}\}$ is a list of starting times corresponding to each of the STL formulas at the internal nodes and predicates at the leaf nodes.
\end{itemize}
\end{definition}

\begin{figure}[t!]
    \centering
    \includegraphics[width=0.45\textwidth]{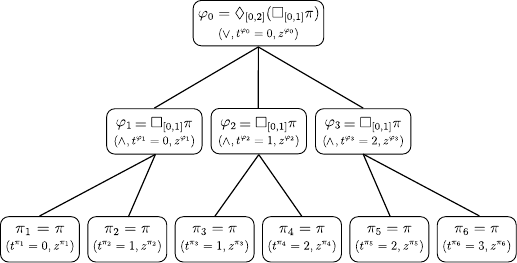}
     \caption{The Logic Tree for $\Diamond_{[0,2]}(\square_{[0,1]} \pi)$, given in Example \ref{exp: example1}.}
\label{fig:example}
\end{figure}

\begin{example}
Consider the specification $\Diamond_{[0,2]}(\square_{[0,1]} \pi)$, whose corresponding LT is shown in Fig. \ref{fig:example}. This tree has 10 nodes, including 6 leaf nodes and 4 internal nodes. The root node has a combination type of disjunction, corresponding to the operator $\Diamond$ in the formula, and three second-level conjunction nodes, corresponding to the operator $\square$ in
the formula.
\label{exp: example1}
\end{example}

To encode the temporal logic constraints represented by an LT into an optimization formulation, \cite{wolff2014optimization} and \cite{raman2014model} propose an MILP, where all variables assigned to the internal nodes $z^{\varphi_i}, \forall i \in \{0, \ldots, n\}$ are continuous variables, and all variables $z^{\pi_i},  \forall i \in \{1, \ldots, |\Pi|\}$ on the leaf nodes are binary variables. For each internal node with a conjunction combination type: $\varphi = \wedge_{i=1}^p \varphi_i$, where $\varphi_i$ is either a formula or a predicate of the child nodes, the following constraints are enforced:
\begin{gather}
z^\varphi \leq z^{\varphi_i}, \ \ i = 1, \ldots , p,  \ \ z^\varphi \geq 1-p+\sum_{i=1}^p z^{\varphi_i}
\label{eqn:tree1}    
\end{gather}
Similarly, for each internal node with a disjunction combination type: $\varphi = \vee_{i=1}^q \varphi_i$ , the following constraints are applied:
\begin{gather}
z^\varphi \geq z^{\varphi_i}, \ \ i = 1, \ldots , q, \ \ z^\varphi \leq \sum_{i=1}^q z^{\varphi_i}
\label{eqn:tree2}  
\end{gather}
On the root node, $z^{\varphi_0}=1$ must hold to satisfy the STL specification. Notably, although $z^{\varphi_i}$ are continuous variables, constraints \eqref{eqn:tree1} and \eqref{eqn:tree2} ensure that $z^{\varphi_i}$ remain binary as long as $z^{\pi_i}$ take binary values. 

In summary, the LT-based optimization formulation is to solve planning problems subject to the dynamics constraint \eqref{eqn:dyn} and the temporal logic specification $\varphi$, while minimizing the objective function $ f_{\rm obj}(\boldsymbol{x}, \boldsymbol{u})$ which typically accounts for factors such as energy consumption and control effort. After constructing the LT $T^\varphi$, the optimization can be formulated as:
\begin{align}&\underset{\substack{\boldsymbol{x}_t \in \mathcal{X} \ \boldsymbol{u}_t \in U \\
z^{\pi_i} \in \mathbb{B} \ z^{\varphi_i} \in [0, 1]}} {\text{minimize}} f_{\rm obj}(\boldsymbol{x}, \boldsymbol{u}) \nonumber \\
& \quad \;\; \quad\begin{aligned}
\text{s.t.}& \quad\;\;~\eqref{eqn:dyn}, \ \ \boldsymbol{x}_0 \in \mathcal{X}_0; \\
& \quad\;\;~\eqref{eqn:predicate},\ \ \forall \: \pi \in \Pi; &\\
& \quad\;\;~\eqref{eqn:tree1} ~\eqref{eqn:tree2}, \ \ \forall \:T^{\varphi_i} \in \mathcal N; \\
& \quad\;\;\ z^{\varphi_0}=1
\end{aligned} 
\label{eqn:motion_planning}
\end{align}


\subsection{Branch and Bound}
\label{sec:BB}

The problem formulation proposed in this study is a mixed-binary linear program (MBLP), which is known to be NP-complete \cite{karp2010reducibility}, and Branch and Bound (B\&B) is a well-accepted method to solve MBLPs. For a feasible optimization problem, B\&B converges to the global optimum; otherwise, it provides a certificate of infeasibility. In this section, we briefly introduce B\&B, and refer readers to \cite{conforti2014integer} for a more detailed explanation. 



Consider an MBLP with continuous variables $\boldsymbol{x} \in \mathbb{R}^{n_x}$, binary variables $\boldsymbol{z} \in \mathbb{B}^{n_z}$, and an optimal objective value $LP^*$. B\&B maintains a search tree, where each node corresponds to a linear programming (LP) problem. These LP problems on nodes are created by relaxing some binary variables $\boldsymbol{z}[j]$, $j=\{1,\ldots,n_z \}$ to continuous variables, and imposing bounds on them. The root node of the tree is a linear program $LP_0$ that relaxes all binary variables to continuous variables. 

Each $LP_i$ in the search tree is associated with a lower bound, $\underline{LP_i}$, on its optimal objective value $LP_i^*$. Heuristics are applied to effectively obtain lower bounds. The B\&B algorithm also keeps an incumbent solution, $\overline{LP}$, which is the best objective value found so far. This value also serves as an upper bound on $LP^*$. If no feasible solution has been found up to the current iteration, $\overline{LP}$ is set to $+\infty$. The success of B\&B relies on efficiently pruning the search tree using both upper and lower bounds, which occurs when $\underline{LP_i}$ is a tight lower bound of $LP_i^*$, and $\overline{LP}$ is a tight upper bound of $LP^*$ \cite{conforti2014integer}.

The relaxation gap of B\&B is defined as $G_a = |\overline{LP} - \underline{LP}| / |\overline{LP}|$, where $\underline{LP}$ is the best lower bound among all $\underline{LP_i}$. $G_a$ is used to measure the tightness of bounds and the solver will terminate when $G_a=0$. In particular, the root relaxation gap is defined as $G_r  = |\overline{LP} - \underline{LP_0}| / |\overline{LP}|$.



\section{Proposed Method}
\label{Sec:logic_flow}

\subsection{Logic Network Flow}

In this paper, we propose a new formulation to encode signal temporal logic specifications, called Logic Network Flow (LNF), which provides a tighter convex relaxation compared to LT. The definition of LNF is given as follows.

\begin{definition}
\label{Def: Logic Network Flow}
A Logic Network Flow $\mathcal{F}^{\varphi}$ derived from a specification $\varphi$ is defined as the tuple $(\mathcal{G}, \mathcal{P}, \Pi, \tau)$, where:
\begin{itemize}
    \item $\mathcal{G} = (\mathcal{V}, \mathcal{E})$ is a directed graph with a source vertex $v_{s} \in \mathcal{V}$ and a target vertex $v_{t} \in \mathcal{V}$.
    \item $\Pi = \{\pi_1, \ldots, \pi_{|\Pi|}\}$ is the set of $|\Pi|$ predicates, each associated with a leaf node in the tree $T^\varphi$ (as defined in Def. \ref{Def: Logic Tree}). We define $\boldsymbol{z}^{\pi} \in \mathbb{R}^{|\Pi|}$ as the vector containing the values of all elements in $\Pi$.
    \item $\mathcal{P}$ is a collection of sets, where each set contains up to $n_e$ ($n_e \leq |\Pi|$) predicates that must hold true to traverse a given edge $e \in \mathcal{E}$. Formally, each set is defined as $P_e \coloneqq \{\pi_i | \pi_i \in \Pi , z^{\pi_i} = 1\} \in \mathcal{P}$. 
    \item $\tau = \{t^{\pi_1}, \ldots, t^{\pi_{|\Pi|}}\}$ is a list of starting times corresponding to each predicate in $\Pi$.
\end{itemize}
\end{definition}








\begin{figure}[t!]
    \centering
    \includegraphics[width=0.48\textwidth]{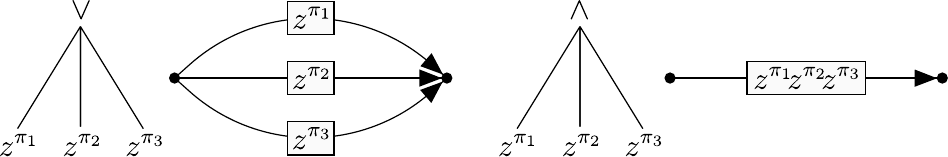}
     \caption{An illustration of the strategy to translate conjunction and disjunction combination types from LTs to LNFs in Algorithm \ref{Algorithm:build_node}.}
\label{fig:example2}
\end{figure} 


Given a vertex $v \in \mathcal V$ in an LNF, let $\mathcal{E}^{\rm in}_{v}$ denote the set of incoming edges to $v$, and $\mathcal{E}^{\rm out}_{v}$ denote the set of outgoing edges from $v$. We present Algorithm \ref{Algorithm:build_node}, a recursive algorithm that translates an LT to an LNF. A similar approach can also be used to construct LNFs directly from STL specifications. To initialize the recursion process in Algorithm \ref{Algorithm:build_node}, we input a source vertex $v_s$, a ``dangling" outgoing edge $e \in \mathcal E^{\rm out}_{v_s}$ without the other end, and an empty predicate set $P_{e} = \varnothing $ associated with the edge $e$. Subsequently, we run the program $\Call{BuildNode}{T^{\varphi_0}, v_{s}, e, P_{e}}$, which returns the target vertex $v_{t}$. Fig. \ref{fig:example4}, \ref{fig:example2} and \ref{fig:example3} show a few examples of converting the LNFs to the LTs.


\begin{algorithm}[t]
	\caption{$\protect \Call{BuildNode}{}$}
        \label{Algorithm:build_node}
         \textbf{Input}: An LT node $T^{\varphi_i}$, a vertex $v$, an outgoing edge $e$ of $v$, and a predicate set $P_e$ for the edge $e$.\\
         \textbf{Output}: A vertex $v$, an outgoing edge $e$ of $v$, and a predicate set $P_e$ for the edge $e$.
	\begin{algorithmic}[1]
    \If{$T^{\varphi_i} = \pi_\psi$ is a leaf node}
        \State Add $\pi_\psi$ to $P_e$.
        \State \textbf{return} $v$, $e$, $P_e$
    \EndIf
    \If{$\circ(T^{\varphi_i}) = \wedge$}
        \For{each child $T^{\varphi_j}$ of $T^{\varphi_i}$}
        \State $v$, $e$, $P_e$ =  \Call{BuildNode}{$T^{\varphi_j}$, $v$, $e$, $P_e$}
        \EndFor
        \State \textbf{return} $v$, $e$, $P_e$
    \EndIf
    \If{$\circ(T^{\varphi_i}) = \vee$}
    \State (\textit{Assume $n$ to be the number of subnodes of $T^{\varphi_i}$})
    \State Duplicate $e$, $P_e$ for $n$ times, denote as $e_j$, $P_{e,j}$, \phantom . \phantom . \phantom . \phantom . \phantom .\phantom . \phantom . where $j=1,\ldots,n$.
    \For{each child $T^{\varphi_j}$ of $T^{\varphi_i}$}
    \State $v$, $e_{o,j}$, $P_{o,j}$ = \Call{BuildNode}{$T^{\varphi_j}$, $v$, $e_j$, $P_{e,j}$}
    \EndFor
    \State Initialize a new vertex $v_{\varphi_i}$, an outgoing edge $e_{\varphi_i}$, \phantom .  \phantom .  \phantom .  \phantom . and a set $P_{e_{\varphi_i}}=\varnothing$ for the edge $e_{\varphi_i}$ \label{new_node} 
    \State Assign $\mathcal{E}_{v_{\varphi_i}}^{\rm in}$ = \{$e_{o,j}$, $j=1,\ldots,n$\}.
    \State \textbf{return} $v_{\varphi_i}$, $e_{\varphi_i}$, $P_{e_{\varphi_i}}$.
    \EndIf
    
	\end{algorithmic} 
\end{algorithm} 


\subsection{Optimization Formulation}
\label{sect:formulation}

\begin{figure}[t!]
    \centering
    \includegraphics[width=0.4\textwidth]{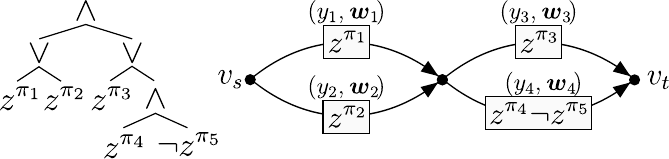}
     \caption{An example of an LT transformed into an LNF by applying Algorithm \ref{Algorithm:build_node}. Each edge in the LNF possesses a binary variable $y_i$ and a vector variable $\boldsymbol{\omega}_i$, given in Example \ref{fig:example3}.}
\label{fig:example3}
\end{figure}

Considering that an LNF is a special instance of Graph-of-Convex-Sets, we propose an optimization formulation for a $\mathcal{F}^{\varphi}$ with a tighter and more compact convex relaxation, similar to the approach in \cite{marcucci2024graphs}. As shown in Fig. \ref{fig:example3}, for each edge $e \in \mathcal{E}$ in an LNF, we associate a binary variable $y_e \in \mathbb{B}$ indicating if this edge is traversed by the in-flow $\boldsymbol{z}^{\pi}$, and a multi-dimensional continuous vector $\boldsymbol{\omega}_e \in [0, 1]^{|\Pi|}$ defined as follows: if the in-flow passes through the edge $e$, we require $\boldsymbol{\omega}_e[i] = 1$ for each predicate $\pi_i \in P_e$, where $\pi_i$ is the $i^{\text{th}}$ predicate in $\Pi$; otherwise, we do not constrain $\boldsymbol{\omega}_e$. To incorporate the connection between $y_e$ and $\boldsymbol{\omega}_e$ into a constraint, we define a selection matrix $\boldsymbol{H}_e \in \mathbb{B}^{n_e \times |\Pi|}$, such that if $\pi_i \in P_e$ is the $i^{\text{th}}$ element in $\Pi$ and the $j^{\text{th}}$ element in $P_e$, we set $\boldsymbol{H}_e[j,i]=1$. All other elements in $\boldsymbol{H}_e$ are set to 0. Let $\boldsymbol{1}_{n_e} \in \mathbb{R}^{n_e}$ be a vector of ones. This constraint can be expressed as:
\begin{equation}
    \boldsymbol{H}_e \boldsymbol{\omega}_e \geq y_e \boldsymbol{1}_{n_e}
\label{eqn:edge}
\end{equation}





For each vertex $v \in \mathcal{V}$ with the input edges $\mathcal{E}_{v}^{\rm in} \subset \mathcal{E}$ and the output edges $\mathcal{E}_{v}^{\rm out} \subset \mathcal{E}$, flow conservation constrains are enforced for both $y_e$ and $\boldsymbol{\omega}_e$:

\begin{equation}
\sum_{e \in \mathcal{E}^{\rm in}_{v}} y_{e} = \sum_{e \in \mathcal{E}^{\rm out}_{v}} y_{e}, \ \ \sum_{e \in \mathcal{E}^{\rm in}_{v}} \boldsymbol{\omega}_{e} = \sum_{e \in \mathcal{E}^{\rm out}_{v}} \boldsymbol{\omega}_{e} 
\label{eqn:vertex}
\end{equation}
In addition, in-flow constraints are imposed to ensure that one unit of flow is injected into the source vertex and the total flow to any vertex does not exceed one unit:
\begin{equation}
\sum_{e \in \mathcal{E}^{\rm out}_{v_s}} y_{e} = 1, \ \ \sum_{e \in \mathcal{E}^{\rm out}_{v_s}} \boldsymbol{\omega}_{e} = \boldsymbol{z}^{\pi}, \ \ \sum_{e \in \mathcal{E}_{v}^{\rm in}} y_{e} \leq 1
\label{eqn:input}
\end{equation}
Notably, if $\mathcal{G}$ is not acyclic, the inequality constraints in \eqref{eqn:input} are essential to prevent cycles in $\mathcal{G}$.
However, since the directed graph $\mathcal{G}$ in an LNF is inherently acyclic, this constraint is not required in our formulation presented in Sec. \ref{sect:complete}. The proof follows directly from topological sorting, which is omitted here for brevity.

\begin{example}
The LNF in Fig. \ref{fig:example3} consists of 4 edges and 3 vertices including $v_s$ and $v_t$. The variables in the LNF are $[y_1, y_2, y_3, y_4]$, $y_i \in [0, 1]$, and  $[\boldsymbol{\omega}_1, \boldsymbol{\omega}_2, \boldsymbol{\omega}_3, \boldsymbol{\omega}_4]$, $\boldsymbol{\omega}_i \in [0, 1]^5$. Edge 1 has a predicate set $P_1=\{z^{\pi_1}\}$, so we enforce $\boldsymbol{\omega}_1[1] \geq y_1$. Similar constraints are applied to Edge 2 and Edge 3. For Edge 4, with the predicate set $P_4=\{z^{\pi_4} \neg z^{\pi_5}\}$, we enforce $\boldsymbol{\omega}_4[4] \geq y_4$ and $ 1-\boldsymbol{\omega}_4[5] \geq y_4$. At the middle vertex, we impose flow conservation constraints $y_1+y_2=y_3+y_4$ and $\boldsymbol{\omega}_1+\boldsymbol{\omega}_2=\boldsymbol{\omega}_3+\boldsymbol{\omega}_4$. For the source vertex $v_s$, the in-flow constraints are $y_1+y_2=1$ and $\boldsymbol{\omega}_1+\boldsymbol{\omega}_2=\boldsymbol{z}^{\pi}$.   
\end{example}

In summary, the LNF formulation ensures that the in-flow $\boldsymbol{z}^{\pi}$ traverses the graph from the source vertex to the target vertex.
If in-flow reaches the target vertex, the specification $\varphi$ is satisfied. In Sec. \ref{sect:complete}, we will explain how the in-flow $\boldsymbol{z}^{\pi}$ couples the LNF with the dynamics in Eqn. \eqref{eqn:dyn}.

\subsection{Incorporating Dynamics as Dynamic Network Flow}

In this study, we abstract the dynamics constraint in~\eqref{eqn:dyn} using a Dynamic Network Flow (DNF) \cite{yu2013multi}. This approach is employed such that, when integrating with LNFs, it preserves the tightness of the convex relaxation from LNFs. We define a set of discrete points $\mathcal{S}$ on a map, and configure a graph connecting these discrete points through offline trajectory optimization. This framework also holds the promise to synthesize with continuous state spaces. 

To build a  DNF, a set of discrete points $\mathcal{S} = \{ \boldsymbol{p}_1, \ldots, \boldsymbol{p}_m \}$ is selected to represent locations that the robots are required to visit based on an STL specification.
A DNF $\mathcal G_D = (\mathcal{E}_D, \mathcal{V}_D)$ is composed of $N\times|\mathcal{S}|$ vertices for a planning trajectory of a horizon $N$. Each vertex $v_{p_i}^t \in \mathcal{V}_D$ has a subscript $p_i$ representing a point $\boldsymbol{p}_i \in \mathcal{S}$ and a superscript $t$ indicating the timestep, i.e., all vertices associated with $\boldsymbol{p}_i$ can be expressed in a sequence of vertices $v_{p_i}^1, v_{p_i}^2, \ldots, v_{p_i}^N$ in the graph.
For each pair of points $(\boldsymbol{p}_i, \boldsymbol{p}_j)$,  edges are connected from $v_{p_i}^t$ to $v_{p_j}^{t+K}$, $\forall t$, if traversing from $\boldsymbol{p}_i$ to $\boldsymbol{p}_j \in \mathcal{S}$ takes $K$ time steps. Edges are also connected from $v_{p_i}^t$ to $v_{p_i}^{t+1}$, representing the robot remaining stationary at timestep $t$. We refer readers to \cite{yu2013multi} for detailed examples of DNFs. 


For each edge $e \in \mathcal{E}_D$, a variable $r_e \in [0, 1]$ is defined to represent the flow carried by the edge. Meanwhile, the flow incurs a cost of $c_e r_e$, which is proportional to the amount of flow $r_e$ with a coefficient $c_e$. Similarly, we impose the flow conservation constraints on all vertices and the in-flow constraints on the source vertex $v_s$:
\begin{equation}
\sum_{e\in \mathcal{E}_{v}^{\rm in}} r_{e} = \sum_{e \in \mathcal{E}^{\rm out}_{v}} r_{e}, \forall v \in \mathcal{V}_D, \ \sum_{e \in \mathcal{E}^{\rm out}_{v_s}} r_{e} = 1
\label{eqn:primitive_dyn}
\end{equation}


\subsection{Complete Formulation}
\label{sect:complete}

In this subsection, we integrate LNFs with DNFs and establish their connections. Recall that $\boldsymbol{z}^{\pi}$, the flow injecting to the LNF, is a vector of binary predicate variables. Binary variable
$z^{\pi_i}$ with starting time $t$ holds true if $(\boldsymbol{a}^{\pi_i})^\top \boldsymbol{p} + b^{\pi_i} \geq 0$, which is equivalent to a flow traversing the vertex $v_{p}^t$ in the DNF. If at least one of the edges input to $v_{p}^t$ is traversed, then $z^{\pi_i} = 1$; if none of these edges is traversed, then $z^{\pi_i} = 0$. Let $\varepsilon$ denote the set of all edges input to the vertex $v_{p}^t$, we arrive at the following constraints connecting $r_e$ and $z^{\pi_i}$:
\begin{equation}
\bigvee\nolimits_{e \in \varepsilon} r_e \Leftrightarrow z^{\pi_i}=1, \quad \bigwedge\nolimits_{e \in \varepsilon} \neg r_e \Leftrightarrow z^{\pi_i}=0
\label{eqn:predicate_new}
\end{equation}
Given this connection between the LNF and the DNF, the complete problem formulation of our approach is expressed as:

\begin{align}
& 
\underset{\substack{r_e \in [0, 1] 
\: z^{\pi_i} \in \mathbb{B} \\ \boldsymbol{\omega}_e \in [0, 1]^{|\Pi|} \ y_e \in \mathbb{B}}}{ \text{minimize}}  \sum_{e \in \mathcal{E}_D} c_e r_e \nonumber \\
& 
\begin{aligned}
\quad\quad\quad\text{s.t.}&  \quad\quad~\eqref{eqn:edge},\ \forall e \in \mathcal{E}; 
\;\; \hspace{-0.05in}\eqref{eqn:vertex},\ \forall v \in \mathcal{V}; \ \\
& \quad\quad~\eqref{eqn:input};  ~\eqref{eqn:primitive_dyn};  ~\eqref{eqn:predicate_new}, \ \forall v \in \mathcal V_D
\end{aligned}
\label{eqn:motion_planning_flow}
\end{align}

\section{Experiments}
\label{Sec:experiments}
In this section, we present two experiments to evaluate the performance of the proposed optimization: (i) optimizing the motions of a team of robots collaboratively navigating a university campus to complete multiple delivery tasks, and (ii) optimizing the motions of several bipeds performing a searching task with bipedal dynamics. Our experiments run on a computer with 12th Gen Intel Core i7-12800H CPU and 16GB memory. All the MIPs are solved using the commercial solver Gurobi 11.0. 
For all task specifications, we solve Eqn.~\eqref{eqn:motion_planning_flow} by converting them into LNFs, and benchmark Eqn.~\eqref{eqn:motion_planning} by converting them into LTs. This benchmark approach, which combines LTs for encoding the logic specifications with DNFs for modeling the dynamic systems, corresponds to the method proposed by \cite{kurtz2021more}.

\begin{figure}[t!]
    \centering
    \includegraphics[width=0.25\textwidth]{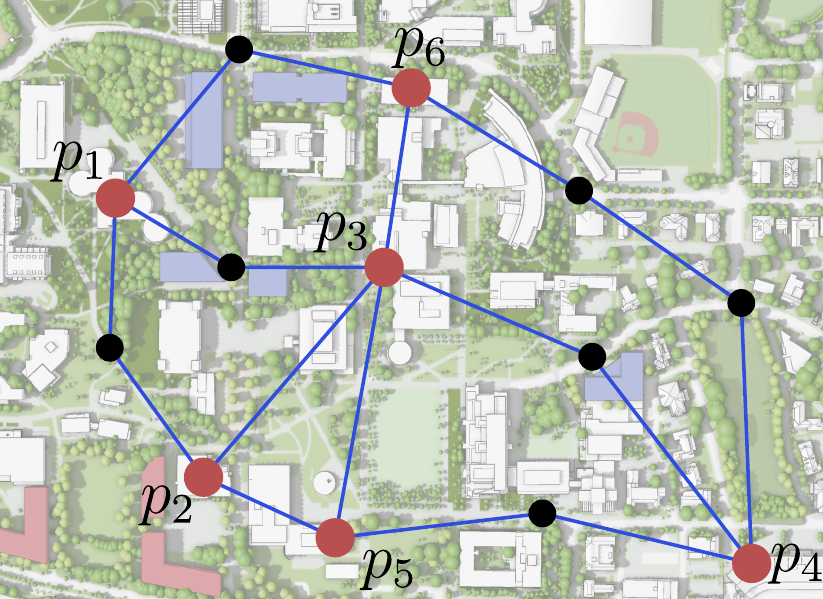}
     \caption{A discretized Georgia Tech map, including 6 sites of interest shown as red dots and 7 interval knots shown as black dots, is used in Sec. \ref{subsec:mutliagent}. It takes each robot $dT$ to travel along each blue line segment divided by black dots in the graph.}
\label{fig:map_exp1}
\end{figure} 

\subsection{Multi-agent Coordination Tasks}
\label{subsec:mutliagent}
We deploy 4 mobile robots on a map with 6 sites of interest ($p_1$ to $p_{6}$) and 7 interval nodes, as shown in Fig. \ref{fig:map_exp1}. The minimum task execution time at any site is $dT=1$ min, with a planning horizon of $N=30$. The robots' motions are modeled as 4 DNFs, each containing $(6+7)\times 30=390$ vertices and 1140 edges. Edge costs are randomly assigned from a uniform distribution $[0, 1]$, representing unexpected factors like wind gusts that affect power consumption, which provides more comprehensive performance evaluation through statistical analysis.


In this example, robots are requested to perform three types of tasks:

\begin{enumerate}
    \item \textit{Delivery tasks}: One robot picks up an item at a site (e.g., $p_1$) during $[t_1,t_2]$ and delivers it to another site (e.g., $p_2$) after $t$ intervals. The specification is:
    \begin{equation*}
        \varphi_{\rm deliver}^{p_1 \rightarrow p_2} = \vee_{i=1}^4(\Diamond_{[t_1, t_2]}(\square_{[0, 2]} z_t^{i, p_1} \wedge \square_{[t, t+2]} z_t^{i, p_2}))
    \end{equation*}
    
    \item \textit{Charging tasks}: Each robot must visit a charging site (e.g., $p_5$) every $10 \sim 20$ minutes. The specification is:
    \begin{equation*}
        \varphi_{\rm charge}^{p_5} = \wedge_{i=1}^4(\Diamond_{[10, 20]}(\square_{[0, 1]} z_t^{i, p_5}))
    \end{equation*}

    \item \textit{Team tasks}: Two robots visit a site (e.g., $p_1$) simultaneously during a interval $[t_1,t_2]$ and remain there for a duration of $t$. The specification is:
    \begin{equation*}
        \varphi_{\rm team}^{p_1} = \Diamond_{[t_1, t_2]}(\vee_{i,j=1, i \neq j}^4(\square_{[0, t]} z_t^{i, p_1} \wedge \square_{[0, t]} z_t^{j, p_1}))
    \end{equation*}
    
\end{enumerate}






We convert the specifications to both LNFs and LTs, and benchmark our optimization formulation in Eqn.~\ref{eqn:motion_planning_flow} against the LT approach in Eqn.~\ref{eqn:motion_planning}. Starting with robots distributed at 4 random vertices, we run 10 tests with different random costs applied to DNFs. We run the tests on three different specifications with increasing complexities: (i) $\varphi_1 = \varphi_{\rm team}^{p_3} \wedge \varphi_{\rm charge}^{p_5}$; (ii) $\varphi_2 = \varphi_{\rm team}^{p_3} \wedge \varphi_{\rm deliver}^{p_4 \rightarrow p_6} \wedge \varphi_{\rm charge}^{p_5}$; (iii) $\varphi_3 = \varphi_{\rm team}^{p_3} \wedge \varphi_{\rm deliver}^{p_4 \rightarrow p_6} \wedge \varphi_{\rm deliver}^{p_2 \rightarrow p_5} \wedge \varphi_{\rm charge}^{p_5}$. 


\begin{figure}[t!]
    \centering
    \includegraphics[width=0.35\textwidth]{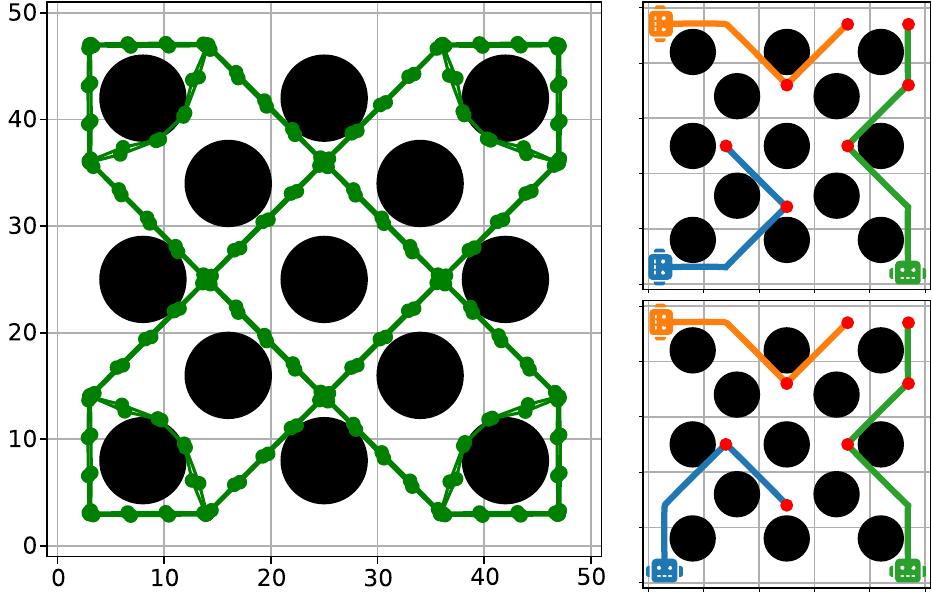}
     \caption{\textit{Left}: The trajectory library for bipedal robot locomotion while navigating around obstacles (shown in black circles). Any position on the map can be locally connected to the vertices (shown by green dots) nearby, and be reached by following the trajectories (shown by green lines). \textit{Right}: Two examples of globally optimal trajectories under two different random costs, both satisfying $\varphi_{\rm search}$. Red dots show the positions required for searching. The orange, blue, and green curves represent the paths taken by each robot.}
\label{fig:map_exp2}
\end{figure} 

\begin{figure*}[t!]
    \centering
    \includegraphics[width=1\textwidth]{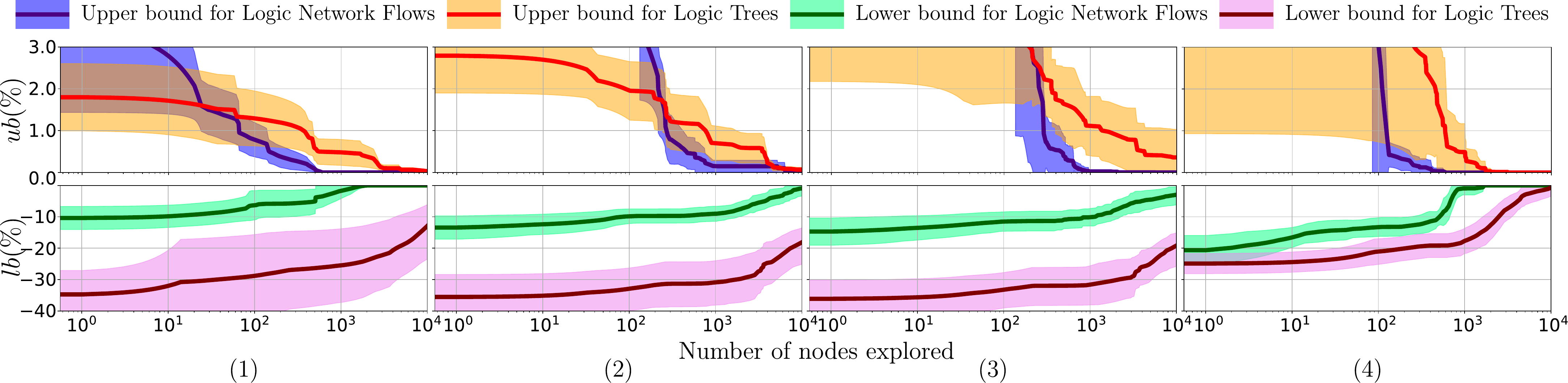}
     \caption{ The comparison between LNFs and LTs in the upper and lower bounds versus the number of nodes explored in the B\&B process for the four tasks given in Sec. \ref{Sec:experiments}.
     From left to right they respectively are 1):  $\varphi_1 = \varphi_{\rm team}^{p_3} \wedge \varphi_{\rm charge}^{p_5}$; 2): $\varphi_2 = \varphi_{\rm team}^{p_3} \wedge \varphi_{\rm deliver}^{p_4 \rightarrow p_6} \wedge \varphi_{\rm charge}^{p_5}$; 3): $\varphi_3 = \varphi_{\rm team}^{p_3} \wedge \varphi_{\rm deliver}^{p_4 \rightarrow p_6} \wedge \varphi_{\rm deliver}^{p_2 \rightarrow p_5} \wedge \varphi_{\rm charge}^{p_5}$; and 4): $\varphi_{\rm search} = \wedge_{j=1}^{7} (\Diamond_{[t_{j,1}, t_{j,2}]}(\vee_{i=1}^3(\square_{[0, 1]} z_t^{i, p_j})))$.
     The four figures on the top shows the upper bounds in percentages, computed by $(\overline{LP} - LP^*)/LP^* \times 100\%$, and the four figures at the bottom shows the lower bounds in percentages, computed by $(\underline{LP} - LP^*)/LP^* \times 100\%$. 
     The curves represent the mean values, and the shaded regions around them represent the variance of the bounds. In general, LNFs discover the same bound by exploring less number of nodes.}
\label{fig:task1}
\end{figure*} 

\renewcommand{\arraystretch}{1.3}
\newcolumntype{C}[1]{>{\centering\let\newline\\\arraybackslash\hspace{0pt}}m{#1}}
\begin{table*}[t!]
\centering
\captionof{table}{Computation results for planning robots' motions in 10 trials}  \label{Tab:results}
\begin{tabular}{@{}l|C{1.3cm}C{1.5cm}|C{1.4cm}C{1.4cm}|C{1.6cm}C{1.7cm}|C{1.4cm}C{1.4cm}}

\multirow{2}{*}{} &
\multicolumn{2}{c|}{$\varphi_{\rm team}^{p_3} \wedge \varphi_{\rm charge}^{p_5}$} &
\multicolumn{2}{c|}{\begin{tabular}[c]{c}$\varphi_{\rm team}^{p_3} \wedge \varphi_{\rm deliver}^{p_4 \rightarrow p_6}$ \\  $\wedge \varphi_{\rm charge}^{p_5}$\end{tabular}} &
\multicolumn{2}{c|}{\begin{tabular}[c]{@{}c@{}}$\varphi_{\rm team}^{p_3} \wedge \varphi_{\rm deliver}^{p_4 \rightarrow p_6} \wedge$ \\ $\varphi_{\rm deliver}^{p_2 \rightarrow p_5} \wedge \varphi_{\rm charge}^{p_5}$\end{tabular}} &
\multicolumn{2}{c}{\begin{tabular}[c]{@{}c@{}}$\wedge_{j=1}^{7} (\Diamond_{[t_{j,1}, t_{j,2}]}$\\ $(\vee_{i=1}^3(\square_{[0, 1]} z_t^{i, p_j})))$\end{tabular}} \\ \cline{2-9}
& Flow & Tree & Flow & Tree& Flow & Tree  & Flow  & Tree  \\ \hline

\# of binary vars & \multicolumn{2}{c|}{124} & \multicolumn{2}{c|}{228}  & \multicolumn{2}{c|}{300} & \multicolumn{2}{c}{633} \\ \hline

\# of cont. vars  & 20682  & 7048  & 42094 & 6989 & 66354  & 6962 & 462177  & 95678 \\ \hline

\# of constr.  & 17417  & 3385  & 39902  & 3739  & 65187      & 4093  & 415807  & 46037  \\ \hline

$G_r$ ($\%$) & \textbf{14 $\pm$ 5} & 51.9 $\pm$ 11.3      & \textbf{15.7 $\pm$ 4.3}  & 54.4 $\pm$ 11.4   & \textbf{16.8 $\pm$ 4.4}    & 55.2 $\pm$ 9.5    & \textbf{25.7 $\pm$ 5.2}      & 56.6 $\pm$ 8.1 \\ \hline

T find opt. (s)   & \textbf{5.5 $\pm$ 1.5}           & 11.5 $\pm$ 8.5     & \textbf{44.0 $\pm$ 16.5}    & 46.0 $\pm$ 41.0   & 138.5 $\pm$ 59.5      & \textbf{78.5 $\pm$ 41.0}  & 58.0 $\pm$ 22.0& \textbf{32.5 $\pm$ 16.0} \\ \hline

\# N find opt.  ($10^3$) & \textbf{0.3 $\pm$ 0.3} & 3.6 $\pm$ 1.7  & \textbf{0.6 $\pm$ 0.3}   & 3.9 $\pm$ 3.6   & \textbf{0.8 $\pm$ 0.2}     & 15.6 $\pm$ 8.0  & \textbf{0.5 $\pm$ 0.2}  & 1.9 $\pm$ 1.1    \\ \hline

T prove opt. (s)  & \textbf{7.8 $\pm$ 2.5} & 53.4 $\pm$ 22.0    & \textbf{231 $\pm$ 104}  & 379 $\pm$ 257 & 1061 $\pm$ 494  & \textbf{897 $\pm$ 540} & \textbf{91 $\pm$ 34}            & 153 $\pm$ 124 \\ \hline

\# N prove opt. ($10^3$)  & \textbf{0.9 $\pm$ 0.6} & 33.1 $\pm$ 19.4   & \textbf{7.3 $\pm$ 2.7} & 79.8 $\pm$ 22.9  & \textbf{10.8 $\pm$ 7.4}  & 140.2 $\pm$ 632.9  & \textbf{1.0 $\pm$ 0.3}& 10.8 $\pm$ 6.1 \\ \hline
\end{tabular}
\end{table*}

Table \ref{Tab:results} shows the number of binary variables, continuous variables, and constraints for each specification, along with the solver results. We report when Gurobi finds the optimal solution (``T find opt.") and how many B\&B nodes are explored (``\# Node find opt."). We also record when optimality is proven (``T prove opt." and ``\# Node prove opt."). Due to high variance in solving times, especially for problems with exponential worst-case time complexity, the bottom four metrics are presented as \textit{median $\pm$ median absolute deviation}. The ``$G_r$" row shows the root relaxation gap between the MBLP relaxation and the global optimum (see Sec. \ref{sec:BB}). Fig. \ref{fig:task1} (1)-(3) display the means and variances of the bounds in relation to the number of nodes explored.


Our results indicate that LNFs outperform LTs in finding better upper and lower bounds, evidenced by the significant reduction in nodes explored to achieve equivalent bound quality. The tighter lower bounds allow B\&B to prune trees more efficiently by detecting unpromising nodes earlier. However, LNFs show computational speed advantages only for simpler specifications ($\varphi_1$ and $\varphi_2$). As logic specifications become more complicated, LNFs introduce more continuous variables and constraints than LTs, resulting in larger convex relaxations at each node of the B\&B tree. For example, the average solving speed per node for LNFs is only 4 times slower than LTs in $\varphi_1$, but 14 times slower in $\varphi_3$. Future work could leverage parallel computing to reduce solving times for large convex programs to improve LNF scalability.


\subsection{Planning Robot Motions to Search over a Map}

In this experiment, three bipedal robots are deployed to search a $50\times50$ meters map containing 13 circular obstacles. The task involves robots visiting several specified sites of interest within a time range given in $\varphi$. To expedite the runtime computation, we utilize an offline-generated trajectory library to link these sites on the map. Those trajectories are generated using linear inverted pendulum dynamics \cite{koolen2012capturability} tailored for bipedal locomotion. The resulting map is given on the left side of Fig. \ref{fig:map_exp2}. When a search position is notified to the robot during runtime, a short trajectory with a horizon of less than 10 seconds is instantaneously planned in less than 100 ms (through IPOPT) to connect the point to the nearest node in the trajectory library.

In this example, 7 positions are chosen from the map, and the swarm is required to visit each position for at least one time. The specification is notated as:
\begin{equation*}
    \varphi_{\rm search} = \wedge_{j=1}^{7} (\Diamond_{[t_{j,1}, t_{j,2}]}(\vee_{i=1}^3(\square_{[0, 1]} z_t^{i, p_j})))
\end{equation*}
We set $dT=4$ sec and the planning horizon $N=45$. Similar to Sec. \ref{subsec:mutliagent}, three DNFs are constructed and random costs are assigned to the edges following a uniform distribution over $[0, 1]$. The varying edge costs reflect different terrain traversability. Two examples of solved paths with each of the 7 search positions visited at least once by one robot are shown on the right side of Fig. \ref{fig:map_exp2}. The two examples differ due to the randomly assigned traversability costs.

The $4^{\rm th}$ column of Table \ref{Tab:results} shows the number of  binary variables, continuous variables, and constraints, along with the computation results explained in Sec. \ref{subsec:mutliagent}. Subfigure (4) in Fig. \ref{fig:task1} displays the mean and variance curves of the bounds versus the number of nodes explored. We observe a similar dramatic decrease in the number of nodes explored to achieve the same upper and lower bounds. However, computational speeds show greater variability because of the problem size. 




\section{Conclusion} 
\label{Sec:conclusion}
This paper proposes LNF, a novel method for encoding STL specifications as MBLPs to enhance the efficiency of the B\&B process. While the initial results are promising, several limitations remain in the current work. First, the findings in this study are empirical: they lack a formal proof to justify the improvement in tightening the bounds. Additionally, LNFs tend to involve more continuous variables and constraints, which increases computation time at each node in the B\&B search tree. In future work, we plan to employ techniques like parallel computing to reduce the computation time per node. Nevertheless, LNFs serve as a valuable alternative to LTs and offer a promising future direction to improving the computational speed of problems with temporal logic specifications.

{
\bibliographystyle{IEEEtran}
\bibliography{references}

\begin{thebibliography}{10}
\providecommand{\url}[1]{#1}
\csname url@samestyle\endcsname
\providecommand{\newblock}{\relax}
\providecommand{\bibinfo}[2]{#2}
\providecommand{\BIBentrySTDinterwordspacing}{\spaceskip=0pt\relax}
\providecommand{\BIBentryALTinterwordstretchfactor}{4}
\providecommand{\BIBentryALTinterwordspacing}{\spaceskip=\fontdimen2\font plus
\BIBentryALTinterwordstretchfactor\fontdimen3\font minus \fontdimen4\font\relax}
\providecommand{\BIBforeignlanguage}[2]{{%
\expandafter\ifx\csname l@#1\endcsname\relax
\typeout{** WARNING: IEEEtran.bst: No hyphenation pattern has been}%
\typeout{** loaded for the language `#1'. Using the pattern for}%
\typeout{** the default language instead.}%
\else
\language=\csname l@#1\endcsname
\fi
#2}}
\providecommand{\BIBdecl}{\relax}
\BIBdecl

\bibitem{plaku2016motion}
E.~Plaku and S.~Karaman, ``Motion planning with temporal-logic specifications: Progress and challenges,'' \emph{AI communications}, vol.~29, no.~1, pp. 151--162, 2016.

\bibitem{zhao2024survey}
Z.~Zhao, S.~Chen, Y.~Ding, Z.~Zhou, S.~Zhang, D.~Xu, and Y.~Zhao, ``A survey of optimization-based task and motion planning: From classical to learning approaches,'' \emph{IEEE/ASME Transactions on Mechatronics}, 2024.

\bibitem{li2021reactive}
S.~Li, D.~Park, Y.~Sung, J.~A. Shah, and N.~Roy, ``Reactive task and motion planning under temporal logic specifications,'' in \emph{2021 IEEE International Conference on Robotics and Automation (ICRA)}.\hskip 1em plus 0.5em minus 0.4em\relax IEEE, 2021, pp. 12\,618--12\,624.

\bibitem{he2015towards}
K.~He, M.~Lahijanian, L.~E. Kavraki, and M.~Y. Vardi, ``Towards manipulation planning with temporal logic specifications,'' in \emph{2015 IEEE International Conference on Robotics and Automation (ICRA)}.\hskip 1em plus 0.5em minus 0.4em\relax IEEE, 2015, pp. 346--352.

\bibitem{shamsah2023integrated}
A.~Shamsah, Z.~Gu, J.~Warnke, S.~Hutchinson, and Y.~Zhao, ``Integrated task and motion planning for safe legged navigation in partially observable environments,'' \emph{IEEE Transactions on Robotics}, 2023.

\bibitem{cardona2023mixed}
G.~A. Cardona, D.~Kamale, and C.-I. Vasile, ``Mixed integer linear programming approach for control synthesis with weighted signal temporal logic,'' in \emph{Proceedings of ACM International Conference on Hybrid Systems: Computation and Control}.\hskip 1em plus 0.5em minus 0.4em\relax Association for Computing Machinery, 2023.

\bibitem{sun2022multi}
D.~Sun, J.~Chen, S.~Mitra, and C.~Fan, ``Multi-agent motion planning from signal temporal logic specifications,'' \emph{IEEE Robotics and Automation Letters}, vol.~7, no.~2, pp. 3451--3458, 2022.

\bibitem{takano2021continuous}
R.~Takano, H.~Oyama, and M.~Yamakita, ``Continuous optimization-based task and motion planning with signal temporal logic specifications for sequential manipulation,'' in \emph{Proceedings of IEEE international conference on robotics and automation}.\hskip 1em plus 0.5em minus 0.4em\relax IEEE, 2021, pp. 8409--8415.

\bibitem{nawaz2024reactive}
F.~Nawaz, S.~Peng, L.~Lindemann, N.~Figueroa, and N.~Matni, ``Reactive temporal logic-based planning and control for interactive robotic tasks,'' \emph{arXiv preprint arXiv:2404.19594}, 2024.

\bibitem{gu2024walking}
Z.~Gu, R.~Guo, W.~Yates, Y.~Chen, Y.~Zhao, and Y.~Zhao, ``Walking-by-logic: Signal temporal logic-guided model predictive control for bipedal locomotion resilient to external perturbations,'' in \emph{2024 IEEE International Conference on Robotics and Automation (ICRA)}.\hskip 1em plus 0.5em minus 0.4em\relax IEEE, 2024, pp. 1121--1127.

\bibitem{gu2024robust}
Z.~Gu, Y.~Zhao, Y.~Chen, R.~Guo, J.~K. Leestma, G.~S. Sawicki, and Y.~Zhao, ``Robust-locomotion-by-logic: Perturbation-resilient bipedal locomotion via signal temporal logic guided model predictive control,'' \emph{arXiv preprint arXiv:2403.15993}, 2024.

\bibitem{nikou2018timed}
A.~Nikou, D.~Boskos, J.~Tumova, and D.~V. Dimarogonas, ``On the timed temporal logic planning of coupled multi-agent systems,'' \emph{Automatica}, vol.~97, pp. 339--345, 2018.

\bibitem{donze2010robust}
A.~Donz{\'e} and O.~Maler, ``Robust satisfaction of temporal logic over real-valued signals,'' in \emph{Proceedings of International Conference on Formal Modeling and Analysis of Timed Systems}.\hskip 1em plus 0.5em minus 0.4em\relax Springer, 2010, pp. 92--106.

\bibitem{mehdipour2019average}
N.~Mehdipour, C.-I. Vasile, and C.~Belta, ``Average-based robustness for continuous-time signal temporal logic,'' in \emph{Proceedings of IEEE Conference on Decision and Control}.\hskip 1em plus 0.5em minus 0.4em\relax IEEE, 2019, pp. 5312--5317.

\bibitem{gilpin2020smooth}
Y.~Gilpin, V.~Kurtz, and H.~Lin, ``A smooth robustness measure of signal temporal logic for symbolic control,'' \emph{IEEE Control Systems Letters}, vol.~5, no.~1, pp. 241--246, 2020.

\bibitem{pant2018fly}
Y.~V. Pant, H.~Abbas, R.~A. Quaye, and R.~Mangharam, ``Fly-by-logic: Control of multi-drone fleets with temporal logic objectives,'' in \emph{Proceedings of ACM/IEEE International Conference on Cyber-Physical Systems}.\hskip 1em plus 0.5em minus 0.4em\relax IEEE, 2018, pp. 186--197.

\bibitem{kurtz2022mixed}
V.~Kurtz and H.~Lin, ``Mixed-integer programming for signal temporal logic with fewer binary variables,'' \emph{arXiv preprint arXiv:2204.06367}, 2022.

\bibitem{marcucci2024graphs}
T.~Marcucci, ``Graphs of convex sets with applications to optimal control and motion planning,'' Ph.D. dissertation, Massachusetts Institute of Technology, 2024.

\bibitem{marcucci2019mixed}
T.~Marcucci and R.~Tedrake, ``Mixed-integer formulations for optimal control of piecewise-affine systems,'' in \emph{Proceedings of ACM International Conference on Hybrid Systems: Computation and Control}, 2019, pp. 230--239.

\bibitem{ahuja1988network}
R.~K. Ahuja, T.~L. Magnanti, and J.~B. Orlin, ``Network flows,'' 1988.

\bibitem{wolff2014optimization}
E.~M. Wolff, U.~Topcu, and R.~M. Murray, ``Optimization-based trajectory generation with linear temporal logic specifications,'' in \emph{Proceedings of IEEE International Conference on Robotics and Automation}.\hskip 1em plus 0.5em minus 0.4em\relax IEEE, 2014, pp. 5319--5325.

\bibitem{raman2014model}
V.~Raman, M.~Maasoumy, and A.~Donz{\'e}, ``Model predictive control from signal temporal logic specifications: A case study,'' in \emph{Proceedings of ACM SIGBED International Workshop on Design, Modeling, and Evaluation of Cyber-Physical Systems}, 2014, pp. 52--55.

\bibitem{yu2013multi}
J.~Yu and S.~M. LaValle, ``Multi-agent path planning and network flow,'' in \emph{Algorithmic Foundations of Robotics X: Proceedings of Workshop on the Algorithmic Foundations of Robotics}.\hskip 1em plus 0.5em minus 0.4em\relax Springer, 2013, pp. 157--173.

\bibitem{belta2019formal}
C.~Belta and S.~Sadraddini, ``Formal methods for control synthesis: An optimization perspective,'' \emph{Annual Review of Control, Robotics, and Autonomous Systems}, vol.~2, no.~1, pp. 115--140, 2019.

\bibitem{leung2023backpropagation}
K.~Leung, N.~Aréchiga, and M.~Pavone, ``Backpropagation through signal temporal logic specifications: Infusing logical structure into gradient-based methods,'' \emph{The International Journal of Robotics Research}, vol.~42, no.~6, pp. 356--370, 2023.

\bibitem{karp2010reducibility}
R.~M. Karp, \emph{Reducibility among combinatorial problems}.\hskip 1em plus 0.5em minus 0.4em\relax Springer, 2010.

\bibitem{conforti2014integer}
M.~Conforti, G.~Cornu{\'e}jols, G.~Zambelli, M.~Conforti, G.~Cornu{\'e}jols, and G.~Zambelli, \emph{Integer programming models}.\hskip 1em plus 0.5em minus 0.4em\relax Springer, 2014.

\bibitem{kurtz2021more}
V.~Kurtz and H.~Lin, ``A more scalable mixed-integer encoding for metric temporal logic,'' \emph{IEEE Control Systems Letters}, vol.~6, pp. 1718--1723, 2021.

\bibitem{koolen2012capturability}
T.~Koolen, T.~De~Boer, J.~Rebula, A.~Goswami, and J.~Pratt, ``Capturability-based analysis and control of legged locomotion, part 1: Theory and application to three simple gait models,'' \emph{The International Journal of Robotics Research}, vol.~31, no.~9, pp. 1094--1113, 2012.

\end{thebibliography}
}

\end{document}